\pdfoutput=1

\documentclass[11pt]{article}

\newif\ifcomment\commenttrue
\newif\ifcomment\commentfalse

\usepackage[a-1b]{pdfx}

\usepackage{framed}
\usepackage{mdwlist}
\usepackage{siunitx}
\usepackage{latexsym}
\usepackage{colortbl}
\usepackage{xcolor}
\usepackage{nicefrac}
\usepackage{booktabs}
\usepackage{fnpct}
\usepackage{amsfonts}
\usepackage[T1]{fontenc}
\usepackage{bold-extra}
\usepackage{amsmath}
\usepackage{amssymb}
\usepackage{bm}
\usepackage{graphicx}
\usepackage{mathtools}
\usepackage{microtype}
\usepackage{multirow}
\usepackage{multicol}
\usepackage{placeins}
\usepackage{xpatch}
\usepackage{latexsym,comment}
\usepackage[normalem]{ulem}

\newcommand*{\missingreference}{{\Huge \colorbox{red}{?reference?}}}
\newcommand*{\missingcitation}{{\Huge \colorbox{red}{?citation?}}}

\makeatletter
\xpatchcmd{\@setref}{\bfseries}{\missingreference}{}{}
\def\@citex[#1]#2{\leavevmode
    \let\@citea\@empty
    \@cite{\@for\@citeb:=#2\do
        {\@citea\def\@citea{,\penalty\@m\ }%
            \edef\@citeb{\expandafter\@firstofone\@citeb\@empty}%
            \if@filesw\immediate\write\@auxout{\string\citation{\@citeb}}\fi
            \@ifundefined{b@\@citeb}{\hbox{\reset@font\missingcitation}%
                \G@refundefinedtrue
                \@latex@warning
                {Citation `\@citeb' on page \thepage \space undefined}}%
            {\@cite@ofmt{\csname b@\@citeb\endcsname}}}}{#1}}
\makeatother

\newcommand{\gem}[1]{\mbox{\textsc{gem}}}
\newcommand{\abr}[1]{\textsc{#1}}

\renewenvironment{quote}
{\list{}{\rightmargin\leftmargin}%
    \item\relax\small\ignorespaces}
{\unskip\unskip\endlist}

\newcommand{\hidetext}[1]{}
\newcommand{\ignore}[1]{}

\ifcomment
    \newcommand{\pinaforecomment}[3]{\colorbox{#1}{\parbox{.8\linewidth}{#2: #3}}}

    \newcommand{\prtodo}[1]{\pinaforecomment{lightblue}{pr}{#1}}
    \newcommand{\prtodoi}[1]{\pinaforecomment{lightblue}{pr}{#1}}
\else
    \newcommand{\pinaforecomment}[3]{}
    \newcommand{\prtodo}[1]{}
    \newcommand{\prtodoi}[1]{}
\fi

\newcommand{\smallurl}[1]{ \begin{tiny}\url{#1}\end{tiny}}

\definecolor{lightblue}{HTML}{3cc7ea}
\definecolor{CUgold}{HTML}{CFB87C}
\definecolor{grey}{rgb}{0.95,0.95,0.95}
\definecolor{ceil}{rgb}{0.57, 0.63, 0.81}
\definecolor{UMDred}{HTML}{ed1c24}
\definecolor{UMDyellow}{HTML}{ffc20e}

\newcommand{\qa}[0]{\abr{qa}}

\newcommand{\nq}[0]{\abr{nq}}

\graphicspath{ {figures/}{auto_fig/} }

\usepackage{acl}

\usepackage{times}
\usepackage{latexsym}
\usepackage{inconsolata}

\usepackage{svg}

\usepackage[T1]{fontenc}

\usepackage[utf8]{inputenc}

\usepackage{microtype}

\usepackage{makecell}
\usepackage{caption}
\usepackage{multirow}
\usepackage{rotating}
\usepackage[most]{tcolorbox}
\usepackage{hyperref}
\usepackage{xcolor}
\usepackage[normalem]{ulem}
\usepackage{booktabs}
\newcommand\hlwrong{\bgroup\markoverwith
  {\textcolor[RGB]{255, 189, 189}{\rule[-.5ex]{2pt}{2.5ex}}}\ULon}

\usepackage{tikz}
\usepackage{subcaption}
\usepackage{supertabular}
\usepackage{collcell}
\definecolor{GradientGreen}{HTML}{30c92a}
\definecolor{UMDYellow}{HTML}{ffc20e}
\newcommand*{\MinNumber}{0}%
\newcommand*{\MidNumber}{100} %
\newcommand*{\MaxNumber}{200}%

\newcommand{\ApplyGradient}[1]{%
        \ifdim #1 pt > \MidNumber pt
            \pgfmathsetmacro{\PercentColor}{max(min(100.0*(#1 - \MidNumber)/(\MaxNumber-\MidNumber),100.0),0.00)} %
            \colorbox{red!\PercentColor!UMDYellow}{#1}
        \else
            \pgfmathsetmacro{\PercentColor}{max(min(100.0*(\MidNumber - #1)/(\MidNumber-\MinNumber),100.0),0.00)} %
            \colorbox{white!\PercentColor!GradientGreen}{#1}
        \fi
}

\newcolumntype{R}{>{\collectcell\ApplyGradient}c<{\endcollectcell}}

\newcommand\textvtt[1]{{\normalfont\fontfamily{cmvtt}\selectfont #1}}

\definecolor{lightgrey}{RGB}{235, 236, 237}
\definecolor{darkgrey}{RGB}{124, 124, 125}

\definecolor{britishracinggreen}{rgb}{0.0, 0.26, 0.15}

\definecolor{lightteal}{RGB}{208,223,226}
\definecolor{teal}{RGB}{69,129,129}

\definecolor{lightorange}{RGB}{252,229,205}
\definecolor{burntorange}{RGB}{207,146,82}

\definecolor{lightpurple}{RGB}{217,210,233}
\definecolor{darkpurple}{RGB}{124,102,179}

\definecolor{lightred}{RGB}{245, 210, 208}
\definecolor{darkred}{RGB}{235, 64, 52}

\definecolor{lightgreen}{RGB}{208, 245, 210}
\definecolor{darkgreen}{RGB}{91, 143, 94}

\definecolor{lightgold}{RGB}{245, 243, 213}
\definecolor{darkgold}{RGB}{145, 136, 12}

\definecolor{lightblue}{RGB}{181, 179, 242}
\definecolor{darkblue}{RGB}{68, 63, 204}

\newcommand{\falseunsure}{{\colorbox{lightred}{\textcolor{darkred}{\texttt{\textbf{\footnotesize False/Unsure}}}}}}
\newcommand{\true}{{\colorbox{lightgreen}{\textcolor{darkgreen}{\texttt{\textbf{\footnotesize True}}}}}}

\newcommand{\instructgpt}{\abr{gpt-\small3.5} }

\newcommand{\sample}{\abr{inference-sample}}

\newtcolorbox[list inside=prompt,auto counter,number within=section]{prompt}[1][]{
    colbacktitle=black!60,
    fonttitle=\small,
    coltitle=white,
    fontupper=\footnotesize,
    boxsep=4pt,
    left=0pt,
    right=0pt,
    top=0pt,
    bottom=0pt,
    boxrule=1pt,
    #1,
}

\title{Pregnant Questions: The Importance of Pragmatic Awareness \\in Maternal Health Question Answering}

\newcommand{\aspace}{\hspace{1em}}
\newcommand{\cs}{$^{\diamondsuit}$}
\newcommand{\biostats}{$^{\spadesuit}$}
\newcommand{\bch}{$^{\clubsuit}$}
\newcommand{\equalcontribute}{$^{*}$}

\author{%
    \textbf{Neha Srikanth}\cs \equalcontribute \aspace 
    \textbf{Rupak Sarkar}\cs \thanks{~\, Equal contribution.} \\
    \textbf{Heran Mane}\biostats \aspace
    \textbf{Elizabeth M. Aparicio}\bch \aspace
    \textbf{Quynh C. Nguyen}\biostats \aspace \\
    \textbf{Rachel Rudinger}\cs \aspace
    \textbf{Jordan Lee Boyd-Graber}\cs \aspace \\
    \cs{}Department of Computer Science, University of Maryland \\ \biostats{}Department of Epidemiology and Biostatistics, University of Maryland \\
    \bch{}Department of Behavioral and Community Health, University of Maryland
     \\
    \texttt{\{nehasrik,rupak\}@umd.edu}
}

\begin{document}
\maketitle
	
	\begin{abstract}
		Questions posed by information-seeking users often contain implicit false or potentially harmful assumptions.
In a high-risk domain such as maternal and infant health, a question-answering system must recognize these pragmatic constraints and go beyond simply answering user questions, examining them in context to respond helpfully.
To achieve this, we study assumptions and implications, or \textit{pragmatic inferences}, made when mothers ask questions about pregnancy and infant care by collecting a dataset of 2,727 inferences from 500 questions across three diverse sources.
We study how health experts naturally address these inferences when writing answers, and  illustrate that informing  existing \qa{} pipelines with pragmatic inferences produces responses that are more complete, mitigating the propagation of harmful beliefs.
	\end{abstract}
	
	\section{Introduction}
	\begin{table*}
\centering
\resizebox{2.08\columnwidth}{!}{
\begin{tabular}{p{0.18\linewidth}p{0.48\linewidth}p{0.6\linewidth}} 
\toprule
& \multicolumn{1}{c}{\textbf{Question}} & \multicolumn{1}{c}{\textbf{Expert-Annotated Pragmatic Inference \& Veracity}} \\ 
\midrule
\textbf{\multirow{3}{*}{\begin{tabular}[c]{@{}l@{}}\abr{rosie}\\(Maternal Health\\QA System) \end{tabular}}} & \texttt{Is it okay for my to color my hair after giving birth?}                & Hair dye chemicals can pass through breast milk from mother to child. (\falseunsure)\\
\arrayrulecolor[rgb]{0.753,0.753,0.753}\cline{2-3}
& \texttt{What is the advantage for not having an epidural during the labor?}    & Avoiding an epidural contributes to a more "natural" and unmedicated birthing experience. (\falseunsure)\\
\cline{2-3}
& \texttt{What cough medicine is appropriate for breastfeeding mothers?} & Some cough medicines can be secreted in breast milk. (\true)\\ 
\arrayrulecolor{black}\midrule
\multirow{3}{*}{\textbf{Reddit}}                                                             & \texttt{Is it safe to lay on my stomach at 28 weeks of pregnancy?} & Sleeping on the stomach while pregnant may have potential risks (\falseunsure) \\
\arrayrulecolor[rgb]{0.753,0.753,0.753}\cline{2-3}
& \texttt{Is it bad to use different bottles/nipples during feedings?} & Using different bottles or nipples for feeding may compromise the baby's latch. (\falseunsure) \\
\cline{2-3}
& \texttt{How can I increase the time between feedings for my 3-month-old baby?} & It may be possible to sleep through the night while still ensuring the baby is fed. (\true)\\ 
\arrayrulecolor{black}\midrule
\multirow{3}{*}{\textbf{Natural Questions}}  & \texttt{When does the fetus begin to develop memory?}  & Fetuses have the ability to form memories. (\true)  \\
\arrayrulecolor[rgb]{0.753,0.753,0.753}\cline{2-3}
& \texttt{What causes a rupture in the amniotic sac?} & There may be ways to prevent early amniotic sac rupture. (\falseunsure)\\
\cline{2-3}
& \texttt{When do the clinical manifestations of an ectopic pregnancy appear?}    & There may be clinical manifestations of an ectopic pregnancy that do not appear early on. (\true)\\
\arrayrulecolor{black}\bottomrule
\end{tabular}
}
\caption{Health experts identify pragmatic inferences from questions from three sources: Reddit, Natural Questions~\cite{kwiatkowski-etal-2019-natural}, and questions asked to our domain-specific \qa{} system, \abr{rosie}~\cite{mane2023practical}. They also determine the veracity of each inference and provide supporting evidence from a trusted web document.}
\label{table:dataset-examples}
\end{table*}

Humans have varying information needs when they ask questions~\cite{taylor1962process}.
Sometimes these needs are easily inferred from the surface form, such as in factoid questions (e.g. \textit{``Who is the 44th president of the United States''}).
However, in a question such as ``\textit{Is there a good non-dairy baby milk I can supplement for my newborn?}'', addressing the underlying false assumption ``\textit{Newborns can safely drink non-dairy milk}'' \textbf{becomes part of satisfying the unexpressed information need}. 
Complete answers to these types of questions must not only address the surface question itself, but also ``question the question'', critically examining its \textit{pragmatic needs}.

These needs become magnified in sensitive domains, such as consumer health or the legal domain.
In these settings, addressing pragmatic needs of questions involves proactively addressing false assumptions or implications in questions to ensure that the asker does not continue holding inaccurate beliefs that they may act on.
For example, a complete answer to the question about non-dairy milk for newborns should address that while non-dairy milk is viable for older babies, newborns and infants need human breast milk or dairy-based formula because they offer complete nutrition.

Language models have been shown to exhibit sycophancy~\cite{sharma2023towards}, sometimes adjusting responses to align with a human user's view.
However, helpful \abr{qa} systems should not only \textit{challenge} false or subjective assumptions in questions~\cite{Kim2022QA2QA} by verifying them against a vetted corpus, but also infer the asker's intent to make sure that its answer satisfactorily addresses their deeper information needs~\cite{taylor1962process}, just as humans do\footnote{``Pregnant'' in our title also refers to its secondary definition, ``full of meaning'' (as in ``a pregnant pause'') alluding to the idea that questions are laden with implicit beliefs.}.
\begin{figure}[t!]
\centering
\includegraphics[scale=0.76]{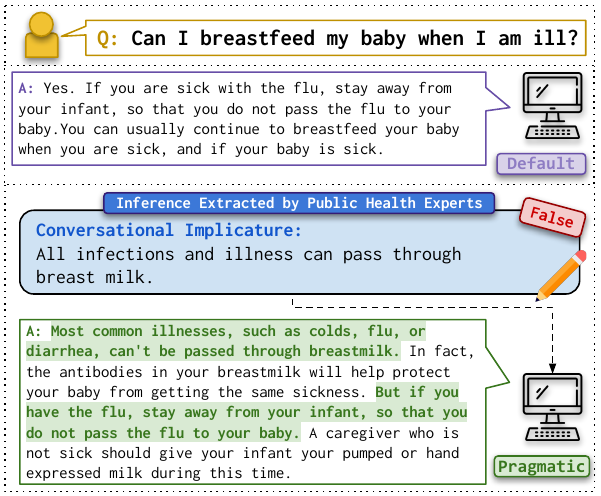}
\caption{We ask public health experts to identify assumptions and implications from questions and find that incorporating them in a \abr{qa} pipeline produces a more complete answer.}
\label{fig:presup-example}
\end{figure}

We construct a dataset\footnote{\scriptsize\url{https://github.com/nehasrikn/pragmatic-inferences-qa}} of 2,727 assumptions and implications in 500 questions (\S\ref{sec:data}) collected from three diverse sources to study (1) how humans embed such assumptions and implications in questions, and (2) the extent to which they are \textit{naturally} addressed in answers written by public health experts.
We then ground assumptions and implications, two primary ways humans embed beliefs in questions, in existing linguistic theory of presuppositions and implicatures respectively (\S\ref{sec:background}).
We refer to presupposition and implicature collectively as \textit{pragmatic inference}.
While recent work has focused on the task of detecting and addressing false \textit{presuppositions} in open-domain \qa{}~\cite{Yu2022CREPEOQ}, we find that false beliefs of question askers are more likely to present as implicatures than presuppositions (\S\ref{sec:ask-and-answer}).
We experiment with \textit{inducing} pragmatic behavior in existing \qa{} pipelines with state-of-the-art retrieval and machine reading models (\S\ref{sec:improve-qa}).
On questions with at least one highly plausible false pragmatic inference, our expert annotators rated responses from our pragmatic \abr{qa} system as more helpful and informative. 

Thus, \qa{} systems of the future must proactively address assumptions and implications in questions as they are increasingly deployed in sensitive domains.

	\section{Collecting Assumptions and Implications in the Wild}
	\label{sec:data}
	In contrast with factoid \qa{}, systems deployed in \textit{sensitive} domains such as consumer health must proactively mitigate harm.
In these settings, correcting false assumptions is \textit{not optional}: systems must provide contextual answers that balance \textit{information completeness} with brevity.

Access to high quality healthcare in the United States vastly differs across socioeconomic backgrounds~\cite{becker2003socioeconomic}.
Such users are often likely to turn to accessible internet resources and---as of late---general purpose chatbots~\cite{palanica2019physicians}.
This motivates us to focus on maternal and infant care, a challenging area of consumer health where patients are concerned with \textit{both} their own physical health as well as the health of their child.

\begin{table}
\centering
\footnotesize
\setlength\tabcolsep{2pt}
\begin{tabular}{lccc} 
\toprule
                    & \begin{tabular}[c]{@{}c@{}}\textbf{Maternal Health}\\\textbf{QA}\end{tabular} & \begin{tabular}[c]{@{}c@{}}\textbf{Reddit}\\\textbf{}\end{tabular} & \begin{tabular}[c]{@{}c@{}}\textbf{Natural}\\\textbf{Questions}\end{tabular}  \\ 
\midrule
\textbf{\# questions}        & 200                                                                                  & 200             & 100                                                                           \\
\textbf{ans. length (\# sent)}    & 3.9                                                                                  & 6.6             & 5.6                                                                           \\
\textbf{\# inferences }      & 1161                                                                                 & 1114            & 452                                                                           \\
\>\>\>\% false/subjective & \textcolor[rgb]{0.125,0.129,0.141}{22.5}                                             & 30.8            & 20.1                                                                          \\
\>\>\>\% true             & 77.5                                                                                 & 69.2            & 79.9                                                                          \\
\bottomrule
\end{tabular}
\caption{Dataset statistics stratified by question source.}
\label{tab:dataset-sizes}
\end{table}

To effectively study and induce pragmatic behavior in \abr{qa} systems, the evaluation questions we choose must reflect real-world experiences and situations for which there may not be a straightforward answer explicitly addressed in a single web document. 
For example, answers to Natural Questions~\cite[\nq{}]{kwiatkowski-etal-2019-natural}---a popular open-domain question-answering dataset---can be found directly in short extracted text snippets from Wikipedia (Table~\ref{table:dataset-examples}). 
In contrast, effectively answering the subjective questions sourced from Reddit requires commonsense reasoning and domain knowledge while identifying the asker's intent.

We carefully construct a dataset of questions from three distributionally distinct sources: a domain-specific \abr{qa} system we design and deploy to pregnant and postpartum participants we recruit~\cite{mane2023practical}, Reddit, and \nq{}.
Then, we introduce an annotation scheme to elicit assumptions and implications from these questions, validate their plausibility, and finally collect supporting evidence to determine their veracity.
Our final dataset contains 2,727 assumptions from 500 evaluation questions (Table~\ref{tab:dataset-sizes}). 
We also include 150 development questions used to train annotators and develop our \abr{qa} systems.

\subsection{Gathering a Diverse Set of Maternal and Infant Health Questions}

\paragraph{Maternal Health \abr{qa} System.}
We source questions come from a maternal and infant health-specific question answering system that we build~\citep{mane2023practical}, henceforth referred to as \abr{rosie}.
Users ask questions pertaining to pregnancy or infant health and are instructed that the \abr{qa} system does not have any personalized knowledge of their individual medical history or pregnancy.

This system operates over a corpus of web documents we construct\footnote{Corpus available upon request. We use \citet{barbaresi-2021-trafilatura} to scrape 408,000 web documents which we split into passages of 100 tokens following \citet{karpukhin-etal-2020-dense}.} from trusted sources including United States governmental and hospital organizations on maternal and infant health, and spans salient topics such as pregnancy and postpartum symptoms, developmental milestones, and infant safety.
Our end-to-end \abr{qa} system, \abr{rosie}, uses a passage retriever and reranker to provide web passages as answers to study participants via a mobile application.
We randomly sample 200 anonymized questions asked to \abr{rosie} for our evaluation set and 50 questions for our development set.

\paragraph{Reddit.} While the questions asked to \abr{rosie} do reflect real-world experiences, they are asked \textit{to an automatic system} and thus tend to include less situational detail or implicit content.
We turn to Reddit\footnote{\texttt{\url{https://www.reddit.com/}}} to capture long-tail questions that are about the diverse set of unique situations a new or expecting parent goes through. 
Table~\ref{table:dataset-examples} highlights some distributional differences between questions from Reddit and other data sources. 
Our questions come from four popular subreddits about maternal and infant health: \texttt{r/BabyBumps}, \texttt{r/breastfeeding}, \texttt{r/NewParents}, \texttt{r/Mommit}, and \texttt{r/beyondthebump} from the \texttt{pushshift}\footnote{\url{https://github.com/Watchful1/PushshiftDumps}} dump.

We develop a series of heuristics as a recall-oriented first step to identify questions with false or subjective assumptions.
We begin by selecting questions where an upvoted comment shows assumption-correcting behavior or where a user invokes their medical expertise, identified by a select list of discourse markers (Appendix~\ref{appendix:reddit}).
Of these, we only retain posts beginning with a ``wh'' word, filtering a few hundred thousand posts down to 2,858 questions. 

As Reddit encourages community participation, many questions are ``community seeking'' as opposed to \textit{information}-seeking. 
To identify information-seeking questions, we use \instructgpt\cite{ouyang2022training} to filter medical questions from non-medical questions (Prompt~\ref{prompt:med-v-non-med}) then manually vet the final set of 297 questions. 
We randomly sample 200 questions for our evaluation dataset and 50 questions for our development set, discarding the rest. 

Titles of Reddit posts are often a hook or a summary of the entire post. 
Using the 50 development questions, we use \instructgpt{} to minimally edit the titular question to include crucial details from the post description, providing a series of exemplars (Prompt~\ref{prompt:reddit-rewrite}).
These rewrites mainly include the age of a newborn or the stage of pregnancy from the description, but sometimes include small situational details that contextualize the question.
Two authors validate all rewrites, keeping the original title wherever both authors agree that the rewrite changed the communicative goal of the asker. 

\paragraph{Natural Questions.} Lastly, we include maternal and infant health questions from \abr{nq} to study pragmatic aspects of factoid-style questions.
We embed all questions in the train set of \abr{nq} using the \texttt{sentence-transformers}~\cite{reimers-2019-sentence-bert} implementation of \texttt{all-mpnet-base-v2}~\cite{DBLP:journals/corr/abs-2004-09297}, including unanswerable questions~\cite{asai-choi-2021-challenges}. 
We identify 2500 answerable questions and 2500 unanswerable questions as maternal health-related by identifying the top 100 nearest neighbors of 50 randomly sampled questions from the development sets of Reddit and \abr{rosie}.\footnote{We tried several different filtering heuristics, including keyword-based detection, but the nearest neighbors approach yielded the most topical questions.}
From this set, we randomly sample 100\footnote{\nq{} questions make up a smaller proportion of our evaluation dataset as we avoid diverting large amounts of annotation resources to factoid-style questions.} questions for our evaluation set and 50 for our development set.
Though obtained with a nearest neighbors approach, these questions greatly differ from those obtained from our previous sources, as they reflect the factoid \abr{qa}-oriented tasks and goals of the original dataset creators (Table~\ref{tab:dataset-sizes}).

\paragraph{Collecting Human Answers from Health Experts.} 
We recruit a team of twenty health experts using Upwork\footnote{\texttt{\url{https://www.upwork.com/}}} to annotate our data including obstetricians and gynecologists (OB/GYNs), nurses, lactation consultants, and public health experts, many of whom have experience with patients. 
In addition, many of these expert annotators are either currently pregnant or postpartum or have been in the past.
We ask a subset of six experts to write helpful and informative long-form answers to all 500 questions in our dataset (Figure~\ref{fig:annotation-instructions}, bottom panel).
While annotators write answers from scratch, they must provide supporting web documents from the same list of verified sources we use to build the corpus for \abr{rosie}.

\subsection{Identifying Assumptions and Implications}

Inferring \textit{possible} assumptions, implications, and asker beliefs from patient questions in our domain are challenging.
In the past, others have extracted assumptions using shallow signals from the surface form of a question~\cite{kim-etal-2021-linguist, parrish-etal-2021-nope}. 
While some assumptions or implications in our dataset can be inferred directly from the question expression, others require deeper domain or experiential knowledge (Table~\ref{table:dataset-examples}). 

Eliciting these assumptions and implications from non-linguists is challenging as existing linguistic frameworks (\S\ref{sec:pragmatic-primer}) are inaccessible or cumbersome for those unfamiliar with the theoretical concepts behind them. 
As such, we operationalize large-scale data collection by asking five annotators from a \textit{different} subset of our expert annotator pool to first write a list of subquestions that an answer to the original question would address (Figure~\ref{fig:annotation-instructions}, top panel). 
Doing so primes annotators to reason about the intent behind a question as well as the information needs of an asker.
Then, we ask them to write a set of sentences reflecting possible beliefs or assumptions that the patient may hold (or, alternatively, beliefs that any complete answer to the question must address).
We emphasize that the assumptions they write can be either medically or scientifically true or false.

Then, we \textit{consolidate} the set of subquestions and human-written assumptions and beliefs into a single set of assumptions and implications using \instructgpt (Prompt~\ref{prompt:question-consolidation}).

\subsection{Annotating Inference Veracity}

Lastly, we ask a new subset of eight expert annotators to annotate whether each assumption and implication in our dataset is medically or scientifically true, false, or subjective and provide a supporting web document from our list of verified sources along with a passage from the document (Figure~\ref{fig:annotation-instructions}, middle panel).

\paragraph{Validation.}
To verify that the assumptions and implications we extract are plausibly inferrable from the question, we recruit an additional pair of health experts, which we refer to as expert \textit{validators}, to rate inferences.\footnote{These expert validators were not a part of our dataset construction.}
We sample 100 assumptions and implications judged as false or subjective, and 100 true inferences and ask our expert validators to rate the plausibility of an inference on a 1--5 Likert scale based on how likely the question asker is to believe the assumption or implication. 
Henceforth, we refer to this sample of 200 inferences coming from 152 unique questions as \sample{}.
Both annotators judge the majority of our inferences as plausible, with 80\% and 95\% rated with a score of at least 3 (see Figure~\ref{fig:inference-validation} for the rating scale).
Spearman’s correlation between the two annotators is 0.69. See Appendix~\ref{appendix:validation-details} for more detail.

	\section{Grounding Assumptions and Implications in Linguistic Theory}
	\label{sec:background}
	Assumptions and implications in our dataset map to two well-studied phenomena in linguistic pragmatics: presupposition and implicature~\cite{grice1975logic, stalnaker1977pragmatic}.
We begin with a short primer of both types of \textit{pragmatic inference} (\S~\ref{sec:pragmatic-primer}) and then discuss the implications of both types in a \abr{qa} setting (\S~\ref{sec:qa-implications}).

\subsection{Two Types of Pragmatic Inference: Presupposition and Implicature}
\label{sec:pragmatic-primer}

A sentence~\textit{S} is a \textit{pragmatic inference} of a
question~\textit{Q} if, depending on the \textit{context}
and \textit{conversational goals} of discourse
participants~\cite{jeretic-etal-2020-natural}, a human would believe that the asker of~$Q$ believes or
assumes~$S$ to be true.
Henceforth, we refer to the assumptions and implications that we collect in our dataset as \textit{pragmatic inferences}.
We review the two most relevant types of pragmatic inferences:
\textit{presupposition} and \textit{implicature}.

\paragraph{Presupposition.} Presuppositions are implicit assumptions in utterances taken for granted by
discourse participants~\cite{beaver1997presupposition}.
The question \textit{``What vitamins should I stop taking after
  becoming pregnant?''} presupposes \textit{``I was taking vitamins
  before becoming pregnant.''}
Presuppositions can often be detected solely 
by the presence of a lexical or syntactic
trigger~\cite{levinson1983pragmatics}.
In the example above, the word \textit{stop} presupposes that an
activity was already in motion.
We refer to these presuppositions as ``trigger-based''. 

As we observe during the collection of our dataset, domain or world knowledge is often needed to
capture presuppositions in real-world data that are not apparent from lexical or
syntactic cues~\cite{abusch2002lexical}.
For example, the question \textit{``Are multiple ultrasounds dangerous for my baby?''} does not directly result in non-trivial trigger-based presuppositions.
However, the asker of the question presupposes that the effects of an ultrasound are additive and hence asks about whether that additive effective is \textit{dangerous}.

\begin{figure*}[t!]
\centering
\includegraphics[scale=0.32]{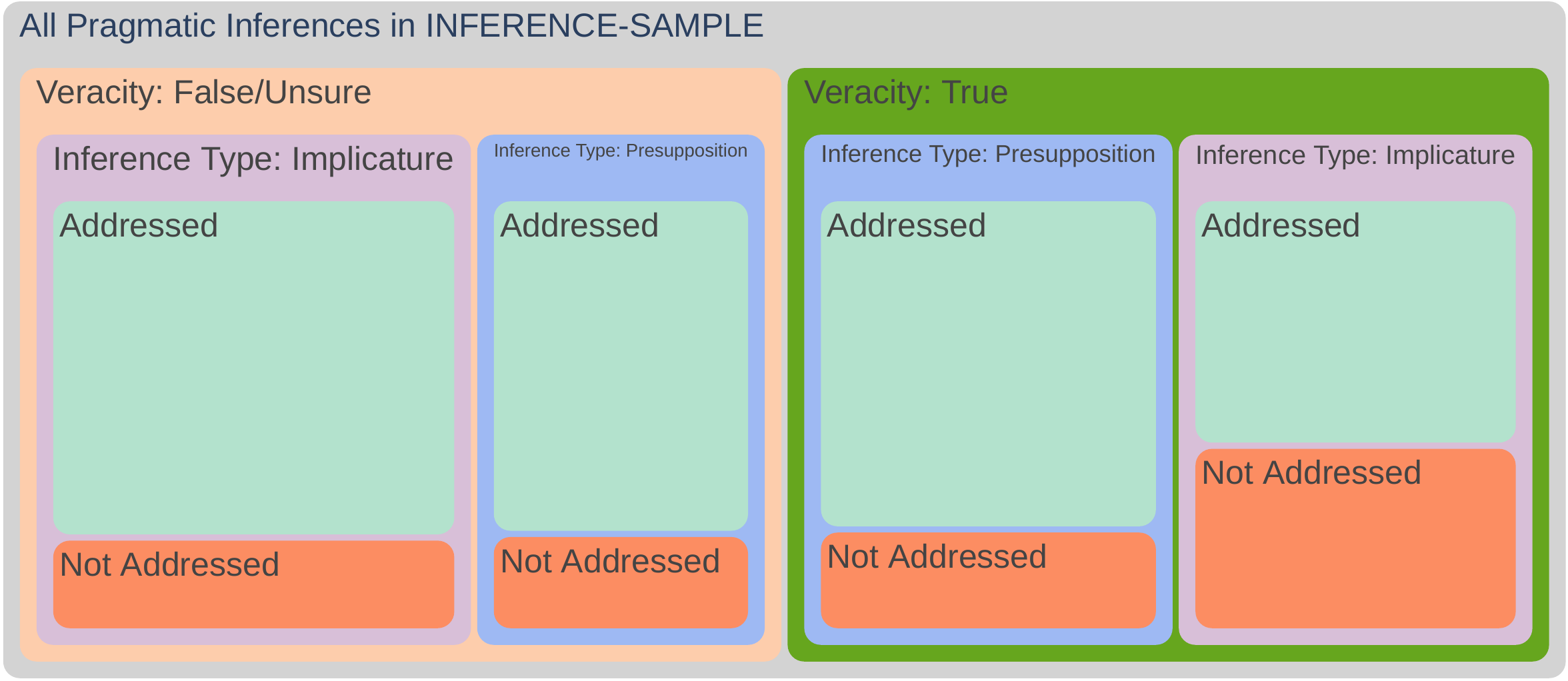}
\caption{Tree-map~\cite{shneiderman1992tree} visualizing the distribution of expert-annotated pragmatic inferences in \sample{} with their veracity, inference type, and whether or not they were addressed or not in the expert-written answer to the question from which they came from. When users make false or subjective inferences, they are more likely to do so as an implicature. Moreover,  when an inference is false, it is more likely to be naturally addressed in an answer by public health experts.}
\label{fig:inference-type-dist}
\end{figure*}

\paragraph{Implicature.} Implicature is a type of pragmatic inference that is
\textit{suggested} by an utterance as opposed to part of its literal
meaning~\cite{grice1975logic}.
Consider the question \textit{``Do most babies fit in newborn clothes?''}
While the speaker understands that newborn clothes fit \textit{some} babies,
their question implies that not all babies fit in newborn clothes.
As we discover, a significant portion of inferences in our dataset are
\textit{implied} from questions rather than presupposed, but detecting
and generating implicatures remain understudied in \abr{nlp}.

Some implicatures are related to lexical items or syntactic
structure of utterances. 
For example, the statement \textit{``These prenatal vitamins are in gumdrop form, \underline{but} are healthy''} implies that gumdrops are usually \emph{not} healthy.
Others are a function of a speaker's intent, beliefs, and other
contextual elements~\cite{zheng-etal-2021-grice}.
While they are a part of the content of an utterance, these implicatures are not \textit{at-issue} (e.g. the main point under discussion~\cite{potts2004logic, koev2018notions}) and are not encoded by the linguistic properties of
a sentence~\cite{ConversationalImplicature}.
Consider the question \textit{``How can you tell the difference between postpartum depression and exhaustion?''}. 
Reasoning about asker belief, we may conclude that they are implying that the two conditions should be treated differently, as one is more serious than the other.

\subsection{Presupposition and Implicature in \qa{}}
\label{sec:qa-implications}

In a natural setting, as we discover, humans embed both presuppositions and implicatures nearly equally in questions (\S\ref{sec:ask-and-answer}).
However, from a linguistic perspective, they represent different
levels of an asker's commitment to the propositional content of the
inference~\cite{peters2016speaker}.
Presuppositions are already a part of an asker's world model. 
In contrast, implicatures are \textit{likely} beliefs that may be negated in an asker's subsequent utterances.
Consider the question \textit{``Is it normal for my baby to move more than usual when closer to due date?''} with both the presupposition \textit{``There are factors that contribute to changes in fetal movement as the due date approaches''} and the implicature \textit{``It may not be necessary to be concerned if there is a significant increase in fetal movement close to the due date.''}
While both are false, the presupposition is stronger, and is clearly in need of addressing in a potential answer.
As illustrated, these distinct phenomena must be dealt with differently when answering a question.

\paragraph{Related Work.} Existing work in pragmatics in \qa{} focuses on open-domain question answering.
\citet{kim-etal-2021-linguist} present the first study of
presuppositions in Google search queries using the Natural Questions
(\abr{nq}) dataset~\cite{kwiatkowski-etal-2019-natural} that are
unanswerable due to false presuppositions.
However, their system only addresses trigger-based presuppositions, overlooking the type of deeper presuppositions present in our dataset derived from world or domain knowledge.
Other work has looked at Google queries with \textit{questionable} assumptions~\cite{Kim2022QA2QA} and false presuppositions in open-domain Reddit questions~\cite{Yu2022CREPEOQ}.
Computational studies of implicature have only focused on specific types, such
as scalar implicature (e.g., some $X$ $\rightarrow$ not all $X$)
\cite{schuster-etal-2020-harnessing, zheng-etal-2021-grice,
  kabbara-cheung-2022-investigating, jeretic-etal-2020-natural}.
As a result of the context induced by our domain, implicatures in our dataset extend beyond scalar implicature.
	
	\section{How do people ask and answer questions?}
	\label{sec:ask-and-answer}
	Before we investigate the behavior of \qa{} systems, we first study
how humans embed pragmatic inferences in their questions
(\S\ref{sec:pragmatic-inference-type}) as well as to what extent they
are \textit{naturally} addressed by human public health experts
(\S\ref{sec:de-novo-support}).

\subsection{Pragmatic Inference Type: Understanding Speaker Commitment}
\label{sec:pragmatic-inference-type}

When users ask questions, how strongly are they committed to the inferences that experts identify in their questions?
Presupposition is a phenomenon based on \textit{mutual} acknowledgment
of facts: when a human makes a presupposition, not only are they
presuming the content of the inference, they are also signaling the
belief that their \textit{interlocutor} (here, a \qa{} system) should
believe it too.

On the other hand, implicatures are a softer way for humans to express
uncertainty.
For example, \textit{``Which immunity injections can I skip for my baby?''} and \textit{``Is it sufficient if my baby takes most immunity injections''} have the same underlying inference (\textit{``It is okay to pick and choose vaccines''}), but is taken for granted in the first (presupposition), whereas loosely suggested in the second (implicature).
We want to distinguish inferences---separating implicatures from presuppositions---in our questions to better characterize \textit{so that we can prioritize addressing stronger false inferences.}

\begin{figure*}[t!]
\centering
\includegraphics[width=\textwidth]{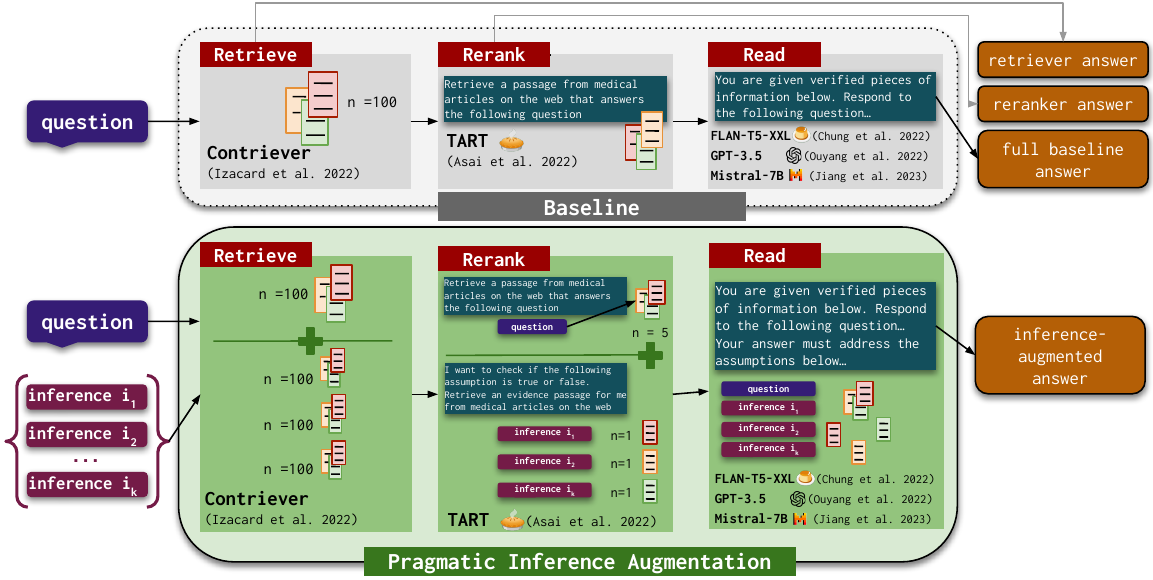}
\caption{Our baseline and pragmatic inference-augmented \abr{qa} pipelines. We experiment with retrieval, reranking, and reading stages and a variety of instruction-tuned and prompt-based models.}
\label{fig:model-pipeline}
\end{figure*}

\paragraph{Annotation Framework.}
A pair of authors independently annotate all inferences in \sample{}
as a presupposition or implicature by first determining whether it is a
proposition about the world that the asker believes to be true,
without which the question would not be felicitous (presupposition) or
whether it involves deriving asker belief through communicative
principles (implicature).
Between authors, Cohen's kappa is $\kappa=0.85$, indicating strong agreement. 
Author annotators adjudicated the final inference type (see Figure~\ref{fig:inference-type-dist} for overall distribution), but individual annotator labels and adjudication rationales are preserved as a part of our dataset.

\paragraph{Findings.}
Presuppositions and implicatures are balanced in \sample{}
(Figure~\ref{fig:inference-type-dist}), with a slight majority of
inferences as implicatures, indicating that many inferences that
health expert annotators identify are more subtle.
When an inference is true, it is almost equally likely to be a presupposition or an implicature. 
However, when users make false or subjective (veracity marked ``Unsure'') inferences, \textbf{they are more likely to do so via implicature} (Figure~\ref{fig:inference-type-dist}).
Past work has looked into generating and verifying
\textit{presuppositions} in open-domain \qa{}, but identifying and addressing \textit{implicatures} in an effort to make answers
information-complete remains heavily underexplored.
This finding highlights a key strength of our work: the ensuing context from our specific domain tests the usefulness of pragmatic inference in \abr{qa} by allowing us to extract a greater range of inferences.

In settings that lack such context (e.g. single-turn open-domain \abr{qa}), we are restricted to leveraging lexical or syntactic signals from the surface form of the question~\cite{kim-etal-2021-linguist} since reasoning about asker belief is not possible without other contextual signals.
For example, in the absence of additional context, the question \textit{``Should I push grandparents for flu shot and tdap?''}~may give rise to inferences involving the safety or effectiveness of these vaccines for the elderly.
However, upon learning that this was asked in a web forum by a postpartum mother, we may reason that she believes her infant may be at risk for contracting the flu or other diseases if their grandparents handle them unvaccinated.

\subsection{Addressing Inferences in Expert Answers}
\label{sec:de-novo-support}
When health experts are tasked with \textit{answering} questions, how likely are they to naturally address inferences that users implicitly make?
Studying whether or not answers \textit{naturally} address pragmatic inferences (\S~\ref{sec:de-novo-support}) gives us better insight into the types of inference health experts, and in turn models, should prioritize when answering questions.

\paragraph{Annotation.} We ask two annotators from our expert annotator pool to determine whether each inference in \sample{} is addressed, either implicitly or explicitly, by the human-written answer to its source question. 

\paragraph{Findings.} The majority of inferences in \sample{} are addressed by the human-written answer \textit{naturally} (Figure~\ref{fig:inference-type-dist}).
Importantly, \textbf{when an inference is false, it is more likely to be naturally addressed.}
Moreover, a significant number of \textit{true} inferences are also addressed by an answer, indicating that health experts not only aim to correct false or subject inferences but also prioritize completeness.
This key finding supports one of the main arguments of this work: \qa{} systems \textit{must} address pragmatic inferences in their answers, just as humans do.

        \section{Inducing Pragmatic Behavior in \qa{}}
	\label{sec:improve-qa}
	Inducing pragmatic behavior in \qa{} systems is not straightforward.
Existing systems are not trained to proactively reason about asker beliefs, since many popular \qa{} datasets do not necessitate this type of behavior (e.g. factoid \qa{}).

We experiment with eliciting model answers that address the pragmatic needs of questions, such as refuting false inferences, using the pragmatic inferences in our dataset.
We inject inferences at each stage of the classic \qa{} pipeline: passage retrieval, reranking, and machine reading (\S\ref{sec:qa-system}) and evaluate outputs against expert-written answers with both automatic and human evaluation (\S\ref{sec:qa-eval}).

\subsection{Experimental Setup}
\label{sec:qa-system}

\paragraph{Corpus.} We use the corpus from \citet{mane2023practical} of 408,000 documents from verified web sources on maternal health and infant care and augment the corpus with the sources that our expert annotators found while both writing answers and determining the veracity of inferences.

\paragraph{Baseline Models.} As a baseline system, we use a retrieval, reranking, and reading-based \abr{qa} pipeline.
Contriever~\cite{izacard2022unsupervised}, an unsupervised dense passage retriever, identifies top relevant documents ($n=100$) in our corpus given a question.
Those documents are reranked using \abr{tart}-full~\cite{asai2022task}, a multi-task retrieval system with a cross-encoder architecture (Instruction~\ref{prompt:tart-baseline}). \abr{tart} is instruction-tuned, equipping it with the flexibility to redefine passage relevance for different tasks.
We feed the top five reranked documents to three different reader models: \abr{flan-t\small5-xxl}~\cite[11 billion parameters]{chung2022scaling}, an instruction-tuned, \textit{prompt}-based encoder-decoder model jointly trained on a multiple tasks with a standard answer extraction prompt from~\citet{naturalinstructions} (Instruction~\ref{prompt:flan-t5-reader}), \abr{mistral-\small7B}~\cite{jiang2023mistral} (an open source large language model, Prompt~\ref{prompt:mistral-gpt-reader}), and \instructgpt (Prompt~\ref{prompt:mistral-gpt-reader}).

\paragraph{Augmenting Systems with Pragmatic Inferences.} In addition to retrieving the top 100 passages using the question as input, we retrieve the top 100 passages for each pragmatic inference of the question ($i_1$...$i_k$) as well. 
Then, for each pragmatic inference $i$, we rerank the top 100 passages using a new inference-informed instruction (Instruction~\ref{prompt:tart-inference}) and select the top passage post-reranking.
We augment the top five reranked passages from the question with these $k$ top passages from each pragmatic inference to feed to each reader (Prompt~\ref{prompt:inference-augmented-reader}).
During reading, we prompt \abr{mistral-7B} and \instructgpt{} to address all $k$ assumptions when generating an answer.\footnote{We do not use \abr{flan-t5-xxl} here because it struggled with reading in the baseline setting.}
To keep the same \textit{number} of passages fed to readers in the baseline pipelines as in the inference-augmented pipeline, we add $k$ extra passages to the top five existing ones. 
This ensures that while the volume of information presented to machine readers is the same in both pipelines, the nature of the content differs, allowing us to measure the utility of inference augmentation during retrieval and reranking.
Figure~\ref{fig:model-pipeline} visualizes our baseline and inference-augmented \abr{qa} pipelines.

\begin{table*}
\centering
\small
\setlength\tabcolsep{2.4pt}
\begin{tabular}{lcc|ccc|cc} 
\toprule
 & \multicolumn{2}{c|}{\textbf{No Reader}} & \multicolumn{3}{c|}{\textbf{Baseline Reader}}     & \multicolumn{2}{c}{\textbf{Inference-Augmented}}  \\ 
\midrule
 & \abr{Retrieve}~   & \abr{Rerank}                 & \abr{Flan-T5-XXL}  & \abr{Mistral-7B}   & \instructgpt      & \abr{Mistral-7B}   & \instructgpt                            \\ 
\midrule
\textbf{ROUGE-L (F1)}                                                                & $15.6_{(10.3)}$  & $17.5_{(10.3)}$    & $15.89_{(13.7)}$    & $17.4_{(5.1)}$   & $18.7_{(6.0)}$   & $16.2_{(4.8)}$   & $18.6_{(5.9)}$                         \\
\textbf{ROUGE-L (Recall)}                                                            & $23.0_{(15.7)}$  & $19.5_{(10.9)}$           & $36.9_{(21.9)}$   & $15.2_{(6.8)}$   & $17.4_{(8.1)}$   & $13.2_{(5.7)}$   & $16.6_{(7.6)}$                         \\
\textbf{BLEURT}                                                                      & -$0.72_{(0.34)}$ & -$0.61_{(0.29)}$          & -$0.79_{(0.45)}$ & -$0.38_{(0.22)}$ & -$0.37_{(0.22)}$ & -$0.36_{(0.21)}$ & -$0.38_{(0.22)}$                       \\
\textbf{\abr{QaFactEval}}                                                     &    $0.69_{(1.05)}$     &    $0.76_{(0.80)}$        &       $1.16_{(1.5)}$      &        $1.02_{(0.65)}$      &     $1.15_{(0.73)}$         &      $0.96_{(0.62)}$        &           $1.17_{(0.75)}$                         \\
\textbf{Human (/5)} & -            & -                        & -            & -            &     $4.43$         & -            &      $4.45$                            \\
\bottomrule
\end{tabular}
\caption{Mean and standard deviations of automatic (\abr{rouge}, \abr{bleurt}, \abr{QaFactEval}) and human evaluation metrics per question. We report results for the top retrieved passage and the top reranked passage, and two modes with and without access to human-written assumptions. Inference-augmented models perform competitively with baselines, indicating the promise of inducing pragmatic behavior in \qa{} models to mitigate harm.}
\label{tab:main-qa-table}
\end{table*}

\subsection{Evaluation}
\label{sec:qa-eval}

We evaluate answers from seven pipeline variations against expert answers (Table~\ref{tab:main-qa-table}):
(1) The \textbf{No Reader} baselines consist of the top retrieved passage based on the input question (\abr{retrieve-only}) and the top reranked passage from the input question (\abr{rerank-only}),
(2) three \textbf{Baseline Readers} (\abr{baseline-flan-t5-xxl}, \abr{baseline-mistral-\small7B}, and \abr{baseline-gpt-\small3.5}), and
(3) two \textbf{Inference-Augmented} pipelines \abr{inference-mistral-\small7B}, and \abr{inference-gpt-\small3.5}). 

\paragraph{Automatic Evaluation Metrics.} Three automatic evaluation metrics measure the quality of generated answers: \abr{rouge}~\cite{lin-2004-rouge} (both F1 and recall),\abr{bleurt}~\cite{sellam-etal-2020-bleurt}, and \abr{QaFactEval}~\cite{fabbri-etal-2022-qafacteval}, a more recent QA evaluation metric originally designed to measure the faithfulness of summaries. 
\instructgpt{} scores the strongest according to \abr{QaFactEval}, our main evaluation metric because it---of the three metrics---most closely captures information content.
However, automatic evaluation of generated answers does \textit{not} capture several higher-level semantic and pragmatic aspects of the question.
Thus, we still need experts to validate the answers.

\paragraph{Human Judgments.}
We ask our expert \textit{validators} to score answers from the top-performing baseline and inference-augmented pipeline (\abr{baseline-gpt-\small3.5} and \abr{inference-gpt-\small3.5}, on \abr{QaFactEval} respectively).
For each of the 152 questions in \sample{}, expert validators score both model outputs simultaneously from 1--5 based on completeness (instructions in Figure~\ref{fig:annotation-instructions-qa-eval}). 
Answers typically received a score of 1 when they were off-topic and missing crucial information, a score of 2 when they were topical but still missing crucial information, 3 when containing all essential information to the question, 4 when most information was present for completeness, and a score of 5 when the answer was information complete.
Judging the information completeness of an answer is a subjective task, as reflected by the Spearman rank correlation between their annotations ($\rho=0.34$). 
While the mean score of inference-augmented examples is comparable to baseline answers ($4.43$ vs. $4.45$), annotators rated the inference-augmented answer as equivalent or better than its baseline counterpart in 75\% of questions in \sample{} (see Table~\ref{tab:generation-examples} for examples).

We further focus on annotator preferences on our original motivating population of questions---those with highly plausible, false assumptions. 
Both annotators rate inference-augmented answers higher than the default answers in the subset of questions with at least one \emph{highly plausible}, (plausibility=5) \emph{false} pragmatic inference (Table \ref{tab:highly-plausible-false}). 
We hypothesize that the similar ratings received by the two systems across \emph{all} questions is due to shortcomings in the instruction-following capabilities of LLMs. 
Forcing the reader model to address pragmatic inferences distracts it from answering the question more completely, and does not always result in more helpful answers when the pragmatic inferences are true.
These results illustrate the promise of inducing pragmatic behavior in \qa{} models and represent a lower bound of their performance, as none of the models we experiment with were trained to optimize for addressing assumptions in questions.

	\section{Can inference extraction be automated?}
	\label{sec:generation}
	While pragmatic inferences elicited from health experts are informed by their expertise, they are slow and costly to collect.
Our \qa{} experiments use \textit{human-written inferences} to
establish an upper bound of answer quality with existing models.
However, a fully automatic pragmatic \abr{qa} pipeline must first generate pragmatic inferences relevant to a question and \textit{then}
generate an answer that addresses the subset of false inferences.
As such, we experiment with generating pragmatic inferences with
\instructgpt\cite{ouyang2022training} to understand to what extent automating the process is feasible with existing prompting and in-context learning.

\paragraph{Experimental Setup.} We generate inferences with \instructgpt{} for all questions in \sample{} using seven in-context examples corresponding to 37 different pragmatic inferences, as more
in-context examples yields diminishing
returns~\cite{liu-etal-2022-makes}.
We select pragmatic inferences written by
multiple expert annotators from diverse user questions and
randomly shuffle them to prevent unwanted effects emerging from example order~\cite{si-relability-2022}, including exemplars from all three sources (\abr{rosie}, Reddit, and \nq{}) to capture distributional differences in their pragmatic inferences.
As humans naturally did, we let \instructgpt{} generate \textit{varied} numbers of inferences per
question.

\paragraph{Evaluation: Can \instructgpt generate human-like pragmatic inferences?}
For each inference in \sample{}, a pair of authors annotate whether or not each human-written assumption is semantically equivalent to at least one inference generated by \instructgpt (Prompt~\ref{prompt:inference-generation}) with a Cohen's kappa of $\kappa=0.88$.
Post-adjudication, 63\% of inferences were not present among model generations. 
When stratifying by inference type, 53\% of presuppositions and 71\% of implicatures were not present.
This illustrates that just as they are more difficult to detect, implicatures grounded in domain knowledge are more difficult for language models to generate.

	\section{Conclusion}
	\label{sec:conclusion}
	We show that it is possible to induce pragmatic behavior in \abr{qa} systems to correct latent false assumptions in the sensitive domain of
maternal and infant health.
Next-generation \abr{qa} systems deployed in real-world settings \textit{must} learn to address the pragmatics of user questions.
Though we have shown the viability of \textit{explicitly} inducing pragmatic behavior in models in this work, directions for future work include training retrievers to inherently search for evidence to address pragmatic inferences and readers to reason on top of such evidence to tactfully and effectively challenge user beliefs.

        \section*{Acknowledgments}
	We thank Philip Resnik, Sweta Agarwal, Sathvik Nair, and other members of the University of Maryland CLIP lab for their helpful feedback.
We also thank Ximena Marin Gutierrez, Marina Yue, Michelle Jasczynski, and Amara Channell Doig for their help in the development of \abr{rosie}.
We are grateful for our pool of expert annotators and for users of \abr{rosie} participating in our clinical trial.

The study reported in this paper was supported by research grants from the National Institute on Minority Health and Health Disparities (grant number R01MD016037) and by the National Library of Medicine (grant number R01LM012849). The content is solely the responsibility of the authors and does not necessarily represent the official views of the National Institutes of Health. The funders had no role in study design, data collection and analysis, decision to publish, or preparation of the paper.
	
	\section*{Limitations}
	\paragraph{Data are not multilingual.} 
Although our participants who provide questions come from diverse socioeconomic and racial backgrounds, all of the data we collect are in English. 
In addition, since we require participants to be located in the United States, the questions provided by participants are only reflective of the healthcare needs of English-speaking residents in the United States.

\paragraph{Choice of a single domain.}
While our approach can be generalized to any other domain, all of our data and experiments are confined to a single domain (maternal and infant health). 
We have not validated that our conclusions generalize beyond this particular important domain.
\paragraph{Pragmatics can be annotator-dependent.}
Finally, some degree of pragmatic inference is always dependent on the annotator, and we have not validated that this is consistent across different annotator backgrounds.
	
	\section*{Ethical Considerations}
        \label{sec:ethics}
	\abr{nlp} systems are never a replacement for doctors or clinical expertise, especially in high-stakes settings. 
This work has grown out of collaboration with public health experts to help disseminate medically accurate but \textit{contextual} information to new or expectant mothers with limited access to healthcare.
Upon detection of false or potentially problematic assumptions, patients can then be referred to healthcare providers better able to provide information than current \abr{qa} systems.
All of our data was collected with \abr{irb} approval in consultation with public health professionals.  

	\bibliographystyle{acl_natbib}
	\bibliography{custom}
	
	\clearpage

    \appendix

\begin{figure}[ht!]
\centering
\includegraphics[scale=0.4]{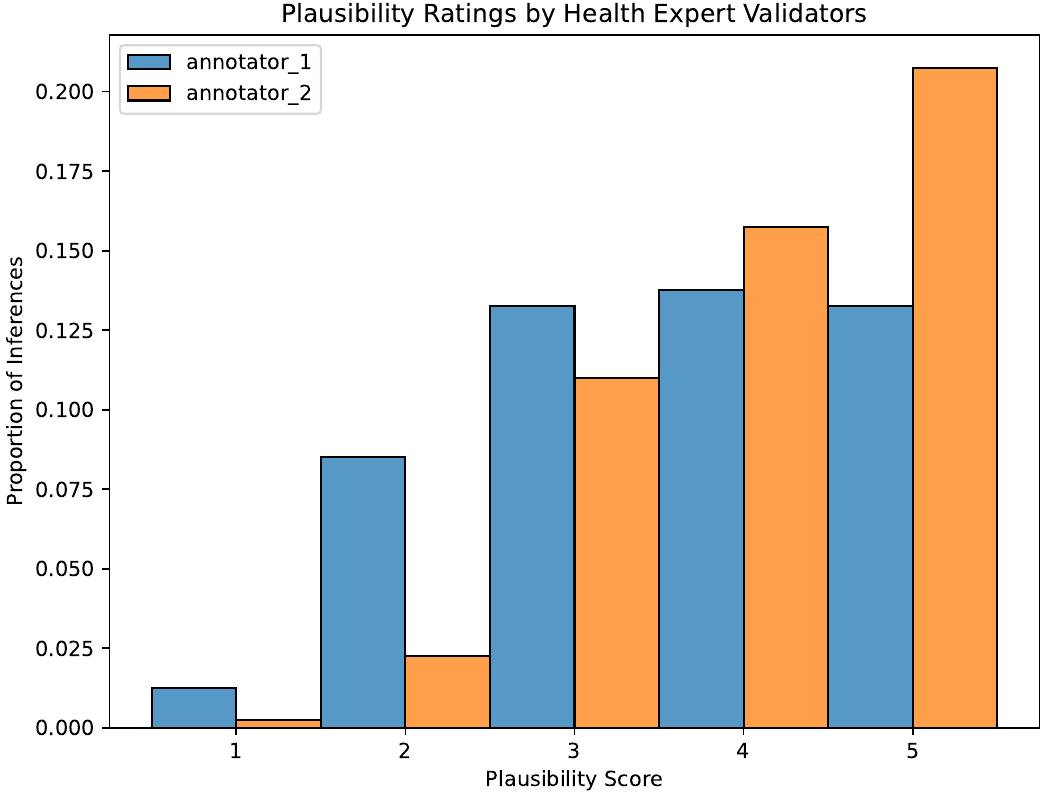}
\caption{Ratings of expert validators of the plausibility of inferences written by health experts in our dataset. The majority of inferences are plausible.}
\label{fig:inference-validation}
\end{figure}

\section{Reddit Question Filtering}
\label{appendix:reddit}

\subsection{Discourse Markers}

{\textvtt{

\paragraph{Assumption Correcting Markers:} 
"however,", "actually,", "as a matter of fact", "in fact", "not true", "despite what you", "on the contrary", "common misconception", "not exactly", "just to clarify", "you're confusing", "correct me if i'm wrong", "correct me if i am wrong", "you're wrong", "we have to remember that", "while that's true", "could be dangerous", "might not be the best thing"

\paragraph{Expertise Invoking Markers:}
"as a doctor", "as a medical professional", "i'm a doctor", "being a doctor", "as a nurse", "i'm a nurse", "i'm a medical professional", "being a nurse"}

\begin{prompt}[title={Prompt \thetcbcounter: Medical vs. Non-Medical Question Identification}, label=prompt:med-v-non-med]
\texttt{\colorbox{lightblue}{Prompt:} You are an expert in maternal and infant health who specializes in finding out whether a question posed by a new or expecting mother is seeking opinion/community participation, or whether it is a medical question. Given a question, you must answer whether it is question seeking medical advice or if it is seeking personal anecdotes and sharing experience. If it's seeking medical advice, answer with ``medical''. Otherwise, answer ``non-medical''. If a question is under-specified, answer with ``non-medical''.}
\end{prompt}

\begin{prompt}[title={Prompt \thetcbcounter: Reddit Question Rewriting}, label=prompt:reddit-rewrite]
\texttt{\colorbox{lightblue}{Prompt:} You will be shown questions about maternal and infant health asked by users. Each question contains a TITLE and DESCRIPTION that elaborates on it, containing details that are both relevant and irrelevant to answering the question. Given a question TITLE and a DESCRIPTION, your task is to incorporate only the relevant details from the DESCRIPTION and rewrite the TITLE into a REWRITE. If there are no relevant details, return the TITLE. As a general rule, keep the rewrite as similar to the original question as possible. The rewrite should be a question in a single sentence. } \\
\texttt{\colorbox{lightblue}{Title:} How to Stop Co-Sleeping}\\
\texttt{\colorbox{lightblue}{Description:} ...}\\
\texttt{\colorbox{lightblue}{Rewrite:} How to wean my 11-month-old out of Co-Sleeping?}
\end{prompt}

\section{Consolidating Subquestions and Assumptions and Implications into Pragmatic Inferences}
\label{appendix:consolidation}

\begin{prompt}[title={Prompt \thetcbcounter: Question Consolidation}, label=prompt:question-consolidation]
\texttt{\colorbox{lightblue}{Prompt:} Given questions asked by new or expecting mothers, your task is to identify the assumptions in them. For this task, you will be given a QUESTION asked by a new or expecting mother, some ASSUMPTIONS (as a list of beliefs or assumptions) in those questions identified by health experts, and some possible SUBQUESTIONS (as a list) that public health experts have identified to have the same information goals as the original question. Given all three of these, your task is to consolidate the SUBQUESTIONS and ASSUMPTIONS into a single, exhaustive list, called INFERENCES. Turning a SUBQUESTION into an inference may involve just turning it into a declarative sentence, or identifying the assumptions made in the SUBQUESTION. Finally, add the INFERENCES to the list of ASSUMPTIONS and remove any duplicates.}
\end{prompt}

\section{Further Details on Inference Validation}\label{appendix:validation-details}

Plausibility scores are the outcome of a three-stage process: (1) a pregnant or new mother holding a belief that is latent while asking a question, (2) a maternal health expert reasoning about these latent beliefs of the mother from the question text, and finally (3) a different expert estimating the likelihood of the beliefs extracted in Step 2 of this process.

The plausibility distribution in Figure~\ref{fig:inference-validation} represents the results of Step 3. It is important to note that the humans involved in each step of the process are completely disjoint, and have little to no information about each other. In 90\% of inferences, the validators agree that the extracted inferences in Step 2 are indeed plausible. The ~10\% of inferences that validators found less plausible is more a reflection of the subjectivity of pragmatic inference in QA in general (see lines 686-690) than our usage of \instructgpt{} a textual transformation tool.

In fact, experts themselves can disagree about answers to questions or background inferences (a finding echoed in \citet{xu-etal-2023-critical}), so we include the plausibility study to reassure audiences that experts do find the inferences in our dataset plausible.

\section{Human Evaluation Details}

\begin{figure}[t!]
\centering
\includegraphics[scale=0.76]{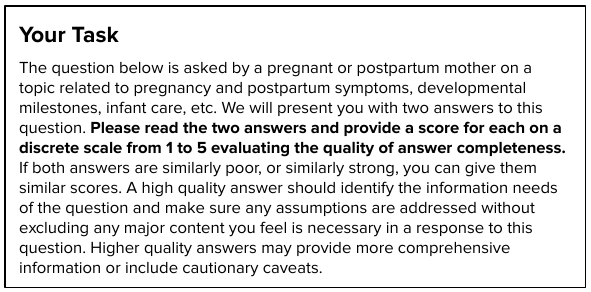}
\caption{Human evaluation instructions provided to two expert annotators.}
\label{fig:annotation-instructions-qa-eval}
\end{figure}

\paragraph{Human Evaluation Annotation Instructions.} See Figure~\ref{fig:annotation-instructions-qa-eval} for annotator instructions.

\paragraph{Annotator preferences on questions with a highly plausible, false assumption.} Human annotators prefer inference-augmented answers for questions with a highly plausible, false assumption. See Table \ref{tab:highly-plausible-false}.
\begin{table}
\centering
\small
\setlength\tabcolsep{2pt}
\begin{tabular}{lcc} 
\toprule
                    & \begin{tabular}[c]{@{}c@{}}\textbf{Mean Likert Score: }\\\textbf{Default Answer}\end{tabular} & \begin{tabular}[c]{@{}c@{}}\textbf{Inference-}\\\textbf{Augmented}\end{tabular}\\ 
\midrule
\textbf{Expert Annotator 1}        & 4.45   & 4.6  \\
\textbf{Expert Annotator 2}    & 4.37   & 4.45    \\
\bottomrule
\end{tabular}
\caption{Human preferences of answers on questions with at least one high-plausibility false assumption.}
\label{tab:highly-plausible-false}
\end{table}

\section{QA Model Prompts}
\label{appendix:qa-prompts}

\subsection{Reranker Instructions}

\begin{prompt}[title={Prompt \thetcbcounter: TART Baseline Instruction}, label=prompt:tart-baseline]
\texttt{Retrieve a passage from medical articles on the web that answers the following question.}
\end{prompt}

\begin{prompt}[title={Prompt \thetcbcounter: TART Pragmatic Inference Instruction}, label=prompt:tart-inference]
\texttt{I want to check if the following assumption is true or false. Retrieve an evidence passage for me from medical articles on the web.}
\end{prompt}
\subsection{Reader Prompts}

\begin{prompt}[title={Prompt \thetcbcounter: Flan-T5 Reader Prompt}, label=prompt:flan-t5-reader]
\texttt{You are an expert in maternal and infant health. You are given a few passages below:
\\
\\
\colorbox{lightblue}{Context:} \{context\} \\
\\
\noindent Using information only from the passages above, respond to the following question with a helpful and complete answer. Use information from multiple passages if necessary to answer the question. Again, you are allowed to only use information from the passages above. \\ \\
\colorbox{lightblue}{Question:} \{question\}
}
\end{prompt}

\begin{prompt}[title={Prompt \thetcbcounter: Mistral/GPT Reader Prompt}, label=prompt:mistral-gpt-reader]
\texttt{You are an expert in maternal and infant health. You are given a few verified pieces of information below: \\
\\
\\
\colorbox{lightblue}{Context:} \{context\} \\
\\
Using evidence only from verified pieces of information above, respond to the following question with a helpful and complete answer. Use information from multiple sources above if necessary to answer the question. In your answer, do not reveal that you are fetching information from these pieces of evidence. Use information only from the verified sources above, and not from any other sources.  \\ \\
\colorbox{lightblue}{Question:} \{question\}\\
\colorbox{lightblue}{Answer:}
}
\end{prompt}

\begin{prompt}[title={Prompt \thetcbcounter: Inference-Augmented Reader Prompt}, label=prompt:inference-augmented-reader]
\texttt{You are an expert in maternal and infant health. You are given a few verified pieces of information below:
\\
\\
\colorbox{lightblue}{Context:} \{context\} \\
\\
Using information only from verified pieces of information above, respond to the following question with a helpful and complete answer. As humans often make assumptions while asking questions, your answer must address assumptions made by the asker listed below.   \\ \\
\colorbox{lightblue}{Assumptions:} \{assumptions\}\\
\\
Use information from multiple sources above if necessary to answer the question below and address the ASSUMPTIONS. In your answer, do not reveal that you are fetching information from these pieces of evidence. Use information only from the verified sources above, and not from any other sources.\\
\colorbox{lightblue}{Question:} \{question\} \\
\colorbox{lightblue}{Answer:}
}
\end{prompt}

\begin{table*}[!ht]
\small
\centering
\resizebox{2.08\columnwidth}{!}{
    \begin{tabular}{p{0.15\linewidth}p{0.45\linewidth}p{0.45\linewidth}}
    \toprule
    \multicolumn{1}{c}{\colorbox{lightpurple}{\textcolor{darkpurple}{\textbf{Question ($Q$)}}}}
    & \multicolumn{1}{c}{\colorbox{lightteal}{\textcolor{teal}{\textbf{Default Answer} ($A_D$)}}} 
    & \multicolumn{1}{c}{\colorbox{lightteal}{\textcolor{teal}{\textbf{Pragmatic Inference-Augmented Answer} ($A_{IA}$)}}} \\ 
    \midrule
    When should I give my baby fever reducing meds after shots? &
    You should give your baby fever reducing medication after shots if they are experiencing discomfort or extreme discomfort due to the fever. However, it is important to note that in most cases, a child does not need to be medicated for a fever unless they are extremely uncomfortable. The most important thing to do when a child has a fever is to make sure they stay hydrated by drinking plenty of fluids. If you are unsure whether your child should receive medication to address the fever or other symptoms, it is recommended to consult with your child's healthcare provider. &
    \textbf{It is not recommended to give your baby fever reducing medication before or immediately after receiving shots. Studies have shown that giving medication prior to a vaccine visit may decrease the child's immune response to the vaccine. Fevers after vaccination are actually a sign that the immune system is responding to the vaccine.} If your baby develops a fever after receiving shots, it is generally not necessary to give them medication unless they are extremely uncomfortable. The most important thing to do when a child has a fever is to make sure they stay hydrated by drinking plenty of fluids. If you are unsure whether your baby should receive medication to address the fever or other symptoms, it is best to consult with your child's healthcare provider.
    \\
    \midrule
    How warm can my bath be while staying safe for my 4-month-old baby? &
    To ensure the safety of your 4-month-old baby during bath time, it is recommended to fill the basin with water that feels warm, but not hot, to the inside of your wrist or elbow. The water temperature should be comfortable for your baby and should not pose a risk of burns or scalds. It is important to check the water temperature before placing your baby in the bath to prevent any potential harm. &
    The safe temperature for a baby may be different from that for an adult. It is important to ensure that the bathwater is warm, but not hot, for a 4-month-old baby. According to the provided information, the water should feel warm to the inside of your wrist or elbow. It is recommended to fill the basin with 2 inches of water at this warm temperature. This is to prevent the baby from getting burned or scalded by hot water. \textbf{It is important to note that babies are not as adaptable as adults to temperature change, and they can lose heat rapidly. Therefore, it is crucial to maintain a safe and comfortable water temperature for the baby during bath time.} \\
    \toprule
    Can pregnancy tests be trusted? &
    According to pregnancy kit manufacturers, most at-home pregnancy tests are 98\% to 99\% accurate when used exactly as instructed. This means that positive results can generally be trusted. However, it is important to note that false negative results can occur if the test is taken too soon. Therefore, it is recommended to wait until a sufficient amount of time has passed since the possible conception before taking a pregnancy test to ensure more accurate results. &
    According to the information provided by pregnancy kit manufacturers, most at-home pregnancy tests are 98\% to 99\% accurate when used exactly as instructed. Positive results from these tests can be trusted. However, it is important to note that false negative results can occur if the test is taken too soon. \textbf{Blood tests, on the other hand, can detect pregnancy sooner and are the only tests that can show specific amounts of HCG in the body...} \\
    \bottomrule
    \end{tabular}
}
\caption{Outputs generated by a standard \abr{qa} system and a \abr{qa} systems informed by pragmatic inferences. Answers from the pragmatic system better explicate their underlying inferences.}
\label{tab:generation-examples}
\end{table*}

\section{Inference Generation Prompt}
\label{appendix:prompt-inference-gen}

\begin{prompt}[title={Prompt \thetcbcounter: Inference Generation Prompt}, label=prompt:inference-generation]
\texttt{When humans ask questions, they often have certain assumptions or implications that are embedded in the questions. These assumptions and implications may be true or false, and they may or may not be present in the surface form of the question. Given a question asked by a new or expecting mother, your task is to identify all relevant assumptions and implications in these questions and write them in a list titled INFERENCES. Each inference under INFERENCES should be an independent and declarative assertion that represents an assumption or an implication that the speaker makes while asking the question. }
\end{prompt}

\begin{figure*}
 \begin{minipage}{\textwidth}
    \begin{subfigure}[b]{\linewidth}
        \centering
        \includegraphics[width=\textwidth]{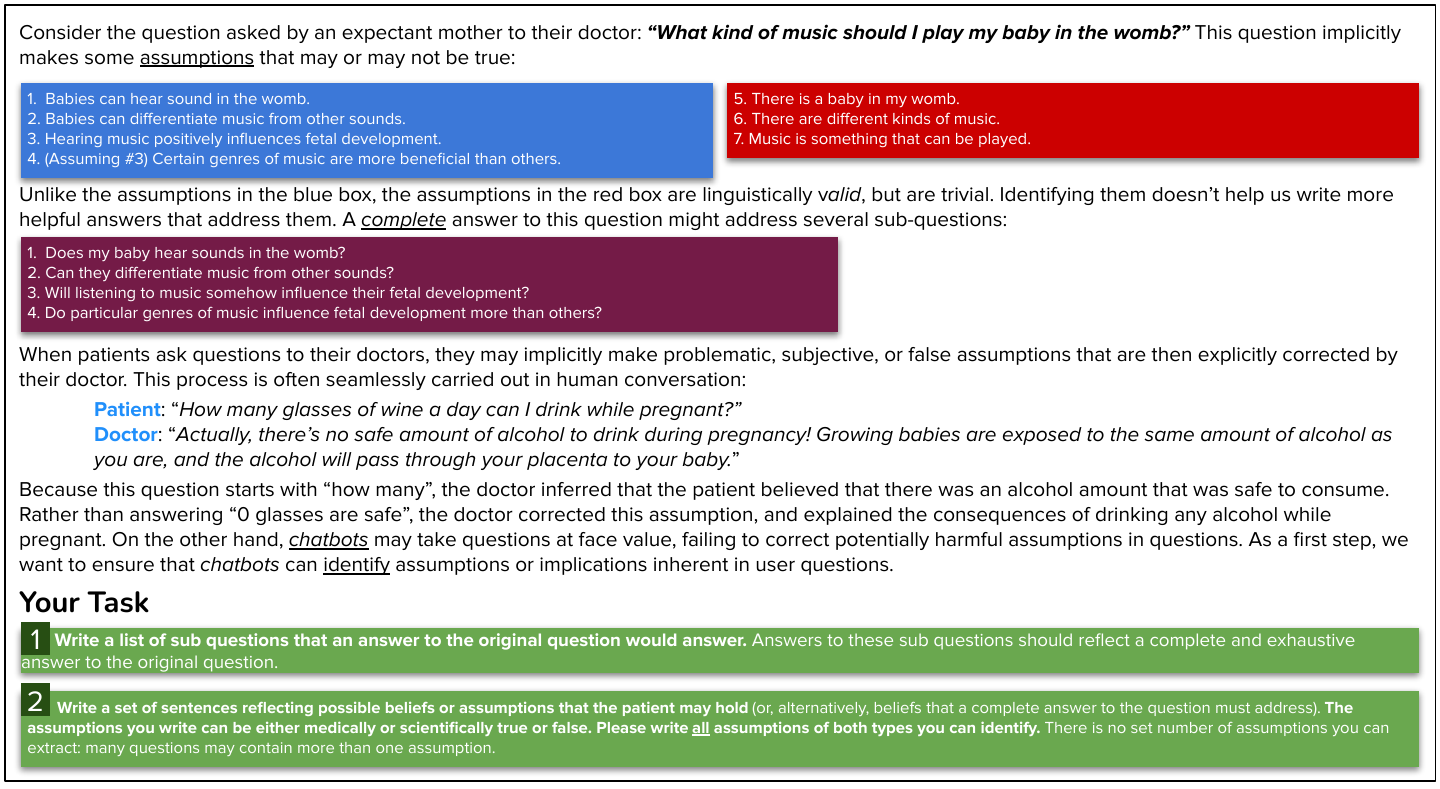}
    \end{subfigure}
    \vspace{4pt}
    \begin{subfigure}[b]{\linewidth}
        \centering
        \includegraphics[width=\textwidth]{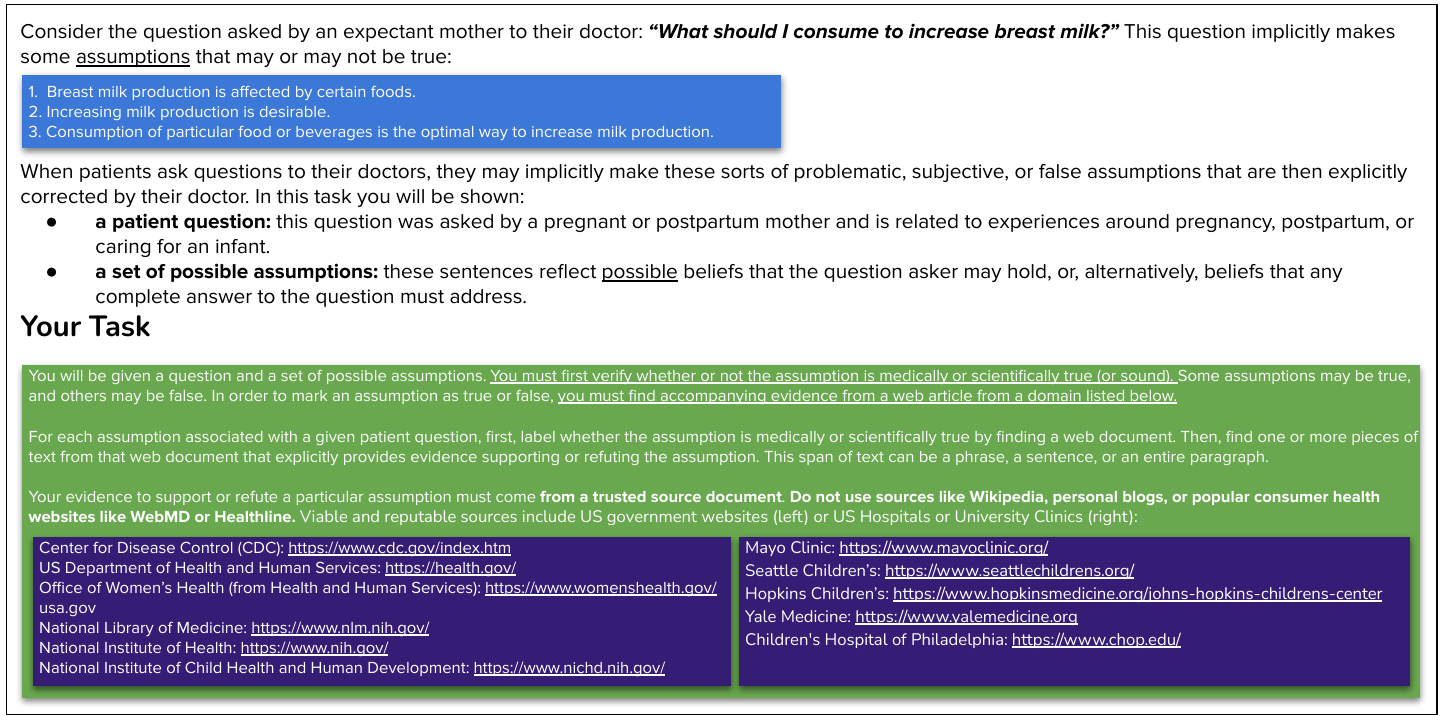}
    \end{subfigure}
    \vspace{4pt}
    \begin{subfigure}[b]{\linewidth}
        \centering
        \includegraphics[width=\textwidth]{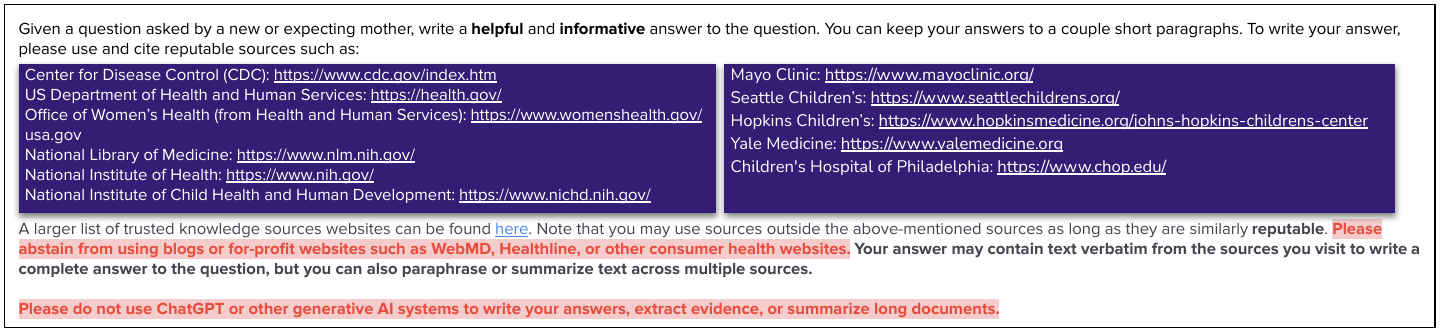}
    \end{subfigure}
  \caption{Instructions given to each annotator for each phase of annotation. First, we show questions to annotators and ask them to write sub questions and the assumptions present (top panel). Then, after passing these outputs to a prompt-based model to extract consolidated inferences, we ask a different set of annotators to \textit{verify} the veracity of the inferences along with supporting evidence (middle panel). Simultaneously, we ask a third set of annotators to write answers to questions without any inference supervision (bottom panel).}
  \label{fig:annotation-instructions}
  \end{minipage}
 \end{figure*}

\end{document}